\documentclass[journal]{IEEEtran}

\usepackage{amsmath,amssymb}
\usepackage{graphicx}
\usepackage{cite}
\usepackage{url}
\usepackage{placeins}
\usepackage{float}
\usepackage{multirow}
\usepackage{array}
\usepackage{booktabs}
\usepackage{tabularx}
\usepackage{longtable}

\title{GLaDiGAtor: Language-Model-Augmented Multi-Relation Graph Learning for Predicting Disease--Gene Associations}

\author{Osman~Onur~Kuzucu~and~Tunca~Do\u{g}an%
\thanks{O.~O.~Kuzucu and T.~Do\u{g}an are with the Biological Data Science Lab, Dept.
of Computer Engineering, Hacettepe University, 06800, Ankara, Turkey.}%
\thanks{T.~Do\u{g}an is also with the Dept.
of Bioinformatics, Graduate School of Health Sciences, Hacettepe University, 06800, Ankara, Turkey.}%
\thanks{T.~Do\u{g}an is also with the Dept.
of Health Informatics, Institute of Informatics, Hacettepe University, 06800, Ankara, Turkey.}%
\thanks{Correspondence to: Tunca Do\u{g}an (tuncadogan@gmail.com).}%
}

\begin{document}

\maketitle

\begin{abstract}
Understanding disease-gene associations is essential for unravelling disease mechanisms and advancing diagnostics and therapeutics. Traditional approaches based on manual curation and literature review are labour-intensive and not scalable, prompting the use of machine learning on large biomedical data. In particular, graph neural networks (GNNs) have shown promise for modelling complex biological relationships. To address limitations in existing models, we propose GLaDiGAtor (Graph Learning-bAsed DIsease--Gene AssociaTiOn pRediction), a novel GNN framework with an encoder--decoder architecture for disease--gene association prediction. GLaDiGAtor constructs a heterogeneous biological graph integrating gene--gene, disease--disease, and gene--disease interactions from curated databases, and enriches each node with contextual features from well-known language models (ProtT5 for protein sequences and BioBERT for disease text). In evaluations, our model achieves superior predictive accuracy and generalisation, outperforming 14 existing methods. Literature-supported case studies confirm the biological relevance of high-confidence novel predictions, highlighting GLaDiGAtor’s potential to discover candidate disease genes. These results underscore the power of graph convolutional networks in biomedical informatics and may ultimately facilitate drug discovery by revealing new gene--disease links. The source code and processed datasets are publicly available at \url{https://github.com/HUBioDataLab/GLaDiGAtor}.
\end{abstract}


\section{Introduction}
Uncovering the molecular basis of human disease is a central challenge in biomedical research and directly informs target identification, patient stratification, and therapeutic development~\cite{ref3,ref5}. In particular, establishing reliable disease--gene associations helps elucidate disease mechanisms, supports variant interpretation in clinical genetics, and enables the prioritization of candidate genes for downstream experimental and therapeutic discovery studies~\cite{ref4}.

Conventional strategies for uncovering disease--gene associations often rely on manual curation and literature review, which, while valuable, are limited by scalability, update frequency, and susceptibility to human bias. To address these limitations, the research community has increasingly adopted machine learning techniques capable of learning from large-scale biological data~\cite{ref7}. In particular, graph-based learning models have proven effective in capturing relational dependencies among biological entities such as genes, proteins, diseases and drugs. From a data-structural and modeling perspective, disease--gene association discovery can be viewed as link prediction on a heterogeneous graph, enabling representation learning for hypothesis generation; similar graph-based formulations also underpin related biomedical tasks such as drug/compound--target interaction (DTI) prediction~\cite{dogan2021druidom} and graph-generative molecular design~\cite{unlu2025}.

Numerous GNN-based methods have emerged in recent years to tackle prediction tasks within biological graphs. For example, SkipGNN~\cite{huang}, HOGCN~\cite{kc}, and ResMGCN~\cite{yin} leverage neighborhood-level message passing for accurate molecular and disease interaction predictions. Foundational models like Graph Convolutional Networks (GCN)~\cite{kipf1}, variational graph autoencoders (VGAEs)~\cite{vgae}, and graph isomorphism networks (GINs)~\cite{xu1} have helped define the theoretical landscape for graph-based learning, while works like JK-Net~\cite{xu2} and MixHop~\cite{abuelhaija} address architectural depth and expressive power. Beyond structural modeling, other approaches have focused on similarity heuristics such as L3~\cite{kovacs} and spectral clustering~\cite{tang}, or utilized unsupervised network embeddings through methods like DeepWalk~\cite{perozzi}, node2vec~\cite{grover}, and struc2vec~\cite{ribeiro}. Despite these advancements, a persistent challenge lies in achieving high generalization, integration of rich biological semantics, and prediction accuracy for novel disease--gene associations.

In this study, we present GLaDiGAtor (Graph Learning bAsed DIsease--Gene AssociaTiOn pRediction), a deep learning framework designed to address these gaps. GLaDiGAtor employs an encoder--decoder architecture to predict associations between diseases and genes using a heterogeneous graph structure. It integrates three types of biological relationships---gene--gene, disease--disease, and gene--disease---from curated biomedical databases. Furthermore, we incorporate contextual node features derived from language models: protein sequences are embedded using ProtT5~\cite{elnaggar}, while disease descriptions are encoded via BioBERT~\cite{ref25}. This multi-relation representation enables GLaDiGAtor to learn enriched feature spaces and infer high-confidence associations. For reproducibility, GLaDiGAtor's source code and datasets are pubicly shared at \url{https://github.com/HUBioDataLab/GLaDiGAtor}.

The primary contributions of this work are as follows:
\begin{itemize}
  \item \textbf{A new GCN-based architecture:} We develop GLaDiGAtor, a new graph convolutional network model using an encoder--decoder paradigm on a heterogeneous gene--disease graph. This design enables the integration of multiple relation types and direct prediction of disease--gene links.
  \item \textbf{Contextual biological embeddings:} We enhance node feature representations with well-known high-performance pretrained embeddings. Each gene/protein node is represented by a ProtT5 sequence embedding, and each disease node by a BioBERT text embedding, providing rich biochemical and semantic context to the model.
  \item \textbf{State-of-the-art performance:} We benchmark GLaDiGAtor against 14 existing methods for disease--gene association prediction, demonstrating consistent improvements in predictive accuracy and generalisation across evaluation datasets.
  \item \textbf{Literature validation of predictions:} We validate GLaDiGAtor’s novel predictions through literature evidence, confirming the biological plausibility of newly predicted gene--disease associations. These case studies illustrate the model’s utility in prioritising candidate genes for diseases, with potential implications for biomedical discovery and drug development.
\end{itemize}

\begin{figure*}[!t]
  \centering
  \includegraphics[width=\textwidth]{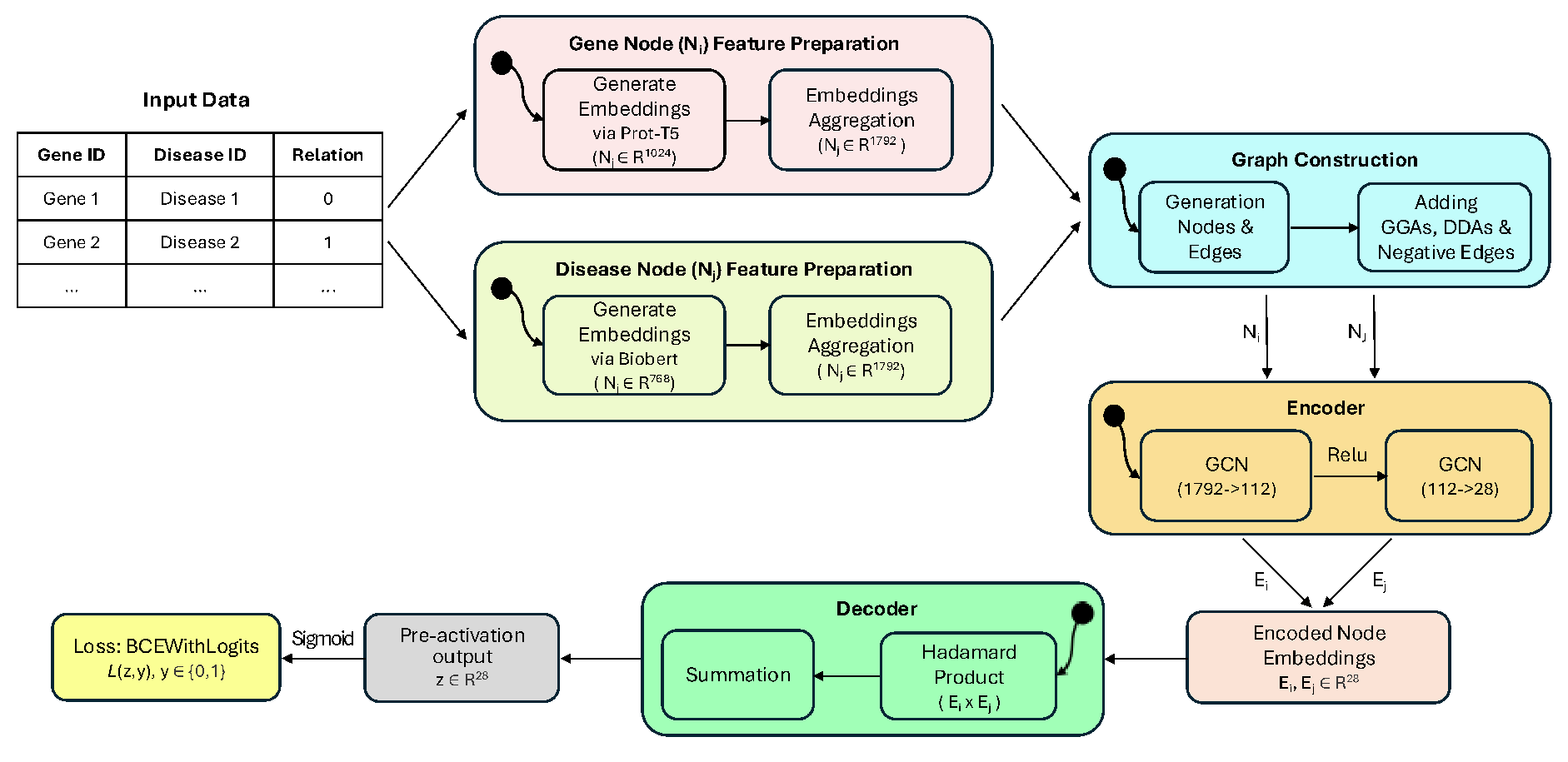}
  \caption{Workflow of the proposed GLaDiGAtor model, illustrating the end-to-end pipeline from input data preparation to encoder--decoder-based graph learning and disease--gene association prediction.}
  \label{fig:gladigator-workflow}
\end{figure*}

\section{Materials and Methods}

\subsection{Heterogeneous Graph Construction}
We constructed a comprehensive heterogeneous graph that includes genes and diseases as nodes, with three types of edges representing known relationships among them. The primary source of disease--gene associations is DisGeNET (v7.0)~\cite{ref1}, a large database of gene--disease links compiled from multiple resources. DisGeNET provides 1,134,942 gene--disease associations (GDAs)~\cite{ref23} with confidence scores and evidence annotations. We obtained two versions of the GDA data from DisGeNET: (a) the complete set via the DisGeNET API, which includes associations aggregated from literature and various databases (each GDA has a score reflecting evidence strength), and (b) a curated subset of high-confidence associations as identified in DisGeNET’s expert-curated section.

In this study, we treat proteins and their coding genes as one entity; therefore, we created a single node for each gene/protein. Naturally, disease--gene associations also become disease--protein relationships.

We integrated protein--protein interactions (PPIs) from the BioGRID database~\cite{oughtred} (also called gene--gene associations---GGAs---here, for the sake of consistency in naming) and disease--disease associations (DDAs) from DisGeNET (computed as disease semantic or phenotypic similarity, e.g., Jaccard indices).

We also considered an external knowledge graph: the Open Graph Benchmark (OGB) ogbl-biokg dataset~\cite{huogb}, from which we extracted protein--disease edges as an alternative source of associations for comparison with methods from the literature (this forms one graph variant, described below). The rest of the relationship types are gathered as described above.

Using these sources, we assembled six graph variants to evaluate the model under different data conditions (summarized from Graph~1 to Graph~6). Each graph follows the same construction pipeline but differs in the origin or filtering of the edges:
\begin{itemize}
  \item \textbf{Graph 1:} DisGeNET (comprehensive data), GDAs filtered to high confidence (score $\ge 0.9$).
  \item \textbf{Graph 2:} DisGeNET, GDAs with score $\ge 0.5$.
  \item \textbf{Graph 3:} DisGeNET, GDAs with score $\ge 0.1$.
  \item \textbf{Graph 4:} DisGeNET, GDAs with score $\ge 0.05$.
  \item \textbf{Graph 5:} DisGeNET, curated GDAs (no score threshold, all curated associations).
  \item \textbf{Graph 6:} OGB (ogbl-biokg), protein--disease associations, in place of DisGeNET GDAs.
\end{itemize}

All graph variants include the same set of gene--gene interactions from BioGRID and disease--disease similarity links from DisGeNET (as available) to provide additional context. Graph statistics are provided in Table~\ref{tab:graph-summary-stats}.

\begin{table}[!t]
\caption{Summary statistics of constructed heterogeneous input graphs.}
\label{tab:graph-summary-stats}
\centering
\renewcommand{\arraystretch}{1.15}
\begin{tabular}{lrrrrr}
\hline
\textbf{Graph} & \textbf{Genes} & \textbf{Diseases} & \textbf{GGA} & \textbf{DDA} & \textbf{GDA} \\
\hline
Graph 1 & 741 & 836 & 3,274 & 462 & 899 \\
Graph 2 & 4,188 & 5,989 & 74,180 & 11,114 & 10,152 \\
Graph 3 & 13,881 & 20,481 & 494,378 & 120,409 & 335,260 \\
Graph 4 & 14,185 & 21,220 & 513,255 & 127,112 & 375,609 \\
Graph 5 & 8,770 & 10,231 & 269,384 & 30,063 & 78,097 \\
Graph 6 & 8,581 & 9,683 & 9,058 & 27,782 & 71,064 \\
\hline
\end{tabular}
\renewcommand{\arraystretch}{1}
\end{table}

The number of nodes and edges varies widely across these graphs, from Graph~1 (the smallest, with under 1k genes and diseases) to Graph~4 (the largest, with $\sim$14k genes, $\sim$21k diseases, and over 1 million total edges across GGA, DDA, GDA). We used consistent node identifiers (NCBI gene IDs and UMLS disease concepts) across all datasets to allow meaningful comparisons.

\subsection{Node Feature Representations}
A key feature of GLaDiGAtor is the integration of pretrained embeddings for both gene and disease nodes, providing biological context beyond the graph topology. We generated protein sequence embeddings for each gene node using ProtT5~\cite{elnaggar}, a transformer model trained on large protein sequence databases. Specifically, for each gene’s protein product, we obtained the ProtT5 embedding (a 1024-dimensional vector) that captures biochemical and evolutionary features of the amino acid sequence. These embeddings encapsulate properties relevant to protein function and interactions.

Similarly, for each disease node, we derived a textual embedding using BioBERT~\cite{ref25}, a BERT-based language model pre-trained on biomedical literature. We input the disease’s name and description into BioBERT to obtain a 768-dimensional vector representation reflecting the semantic context of the disease. This provides the model with an understanding of disease-related information in textual form.

Because the gene and disease embeddings have different dimensionalities (1024 vs. 768), we performed a simple feature alignment to combine them. We zero-padded each vector to a fixed length of 1792 (the sum of 1024 and 768). In practice, we pad the gene vectors with zeros at the end (suffix) and the disease vectors with zeros at the beginning, so that both become 1792-dimensional and can be used in the same feature space. This differential padding scheme was chosen based on preliminary testing of alternatives (as discussed in the ablation study): padding gene and disease embeddings on opposite ends yielded slightly better performance than other alignment strategies. After alignment, each node (whether gene or disease) is represented by a 1792-dimensional initial feature vector. These node features are used as input to the graph neural network.

\subsection{Dataset Finalisation and Partitioning}
Once the graph was constructed, we performed a UniRef protein sequence clusters-based~\cite{suzek} data split to prepare training and evaluation sets. To prevent information leakage due to closely related genes/proteins appearing in both training and test sets, we grouped them by their UniRef50 protein clusters. Proteins within the same cluster (i.e., sharing $\ge 50\%$ sequence identity) were assigned to the same data partition. After enforcing this, we randomly split the gene--disease edges (the target associations) into training, validation, and test sets in an 8:1:1 ratio. The same protocol is used across all graph variants in Table~\ref{tab:graph-summary-stats}. All gene--gene and disease--disease edges were kept available during training (they are considered part of the known network context, not prediction targets) unless explicitly removed in an ablation experiment.

For model supervision, we treated the presence of a gene--disease link as a positive instance. We then generated an equal number of negative instances by sampling gene--disease pairs that are not connected in the graph, ensuring that each negative instance consists of a gene and a disease that do not have a known association. Negative sampling was type-constrained (only gene nodes paired with disease nodes) and one-to-one balanced to match the count of positive edges in each split. This strategy yields a balanced binary classification setup. We also applied degree-aware sampling for negatives to mitigate bias: when drawing negative edges, diseases were weighted by node degree so that high-degree disease nodes were sampled with higher probability. This preserves a realistic difficulty where widely studied diseases have many candidate gene links to reject potentially.

We assembled graphs in NetworkX~\cite{hagberg} for flexible manipulation and integrity checks, then converted them to PyTorch Geometric~\cite{fey} format to enable efficient mini-batch training and GPU-accelerated message passing in GNNs.

\subsection{Model Architecture}
We model a heterogeneous, undirected graph $G=(V,E)$ with three relation types $R=\{\mathrm{GG},\mathrm{DD},\mathrm{GD}\}$. Each node has a $1\times 1792$ feature vector (aligned gene/protein and disease embeddings). GLaDiGAtor uses a GCN encoder to obtain $d$-dimensional node embeddings ($d=28$) and a bilinear decoder to score gene--disease links. The end-to-end pipeline follows the draft: relation-wise normalization with self-loops, convex mixing of relations, two linear GCN transforms with ReLU, and a bilinear logit scored with a sigmoid for probabilities.

\subsubsection{Encoder}
The encoder of GLaDiGAtor is a two-layer Graph Convolutional Network (GCN) that transforms high-dimensional node features into compact latent embeddings. Let the input feature matrix be given by
\begin{equation}
X \in \mathbb{R}^{|V|\times 1792}
\label{eq:input-feature-dim}
\end{equation}
where $|V|$ is the number of nodes and each node is described by 1792 features. We denote $H^{(0)}=X$.

The first GCN layer aggregates information from neighboring nodes and applies a linear transformation followed by a non-linear activation:
\begin{equation}
H^{(1)}=\sigma\!\left(\tilde{A}H^{(0)}W^{(0)}\right),
\label{eq:gcn-layer1}
\end{equation}

where $\tilde{A}$ is the adjacency matrix of the constructed graph, $W^{(0)}\in\mathbb{R}^{1792\times 112}$ is a trainable weight matrix, and $\sigma(\cdot)$ denotes the ReLU activation function.

The second GCN layer further refines the node representations:
\begin{equation}
Z=\sigma\!\left(\tilde{A}H^{(1)}W^{(1)}\right),
\label{eq:gcn-layer2}
\end{equation}

with $W^{(1)}\in\mathbb{R}^{112\times 28}$. The output $Z\in\mathbb{R}^{|V|\times 28}$ provides a 28-dimensional latent embedding for each node.

To mitigate overfitting, dropout with fixed rate $p$ is applied after each GCN layer. The choice of two layers and a relatively low-dimensional final embedding ($d=28$) was guided by validation experiments, balancing expressiveness and generalization.

\subsubsection{Decoder}
The decoder of GLaDiGAtor takes the latent node embeddings produced by the encoder and generates predictions about potential relationships between node pairs. Specifically, it operates on the 28-dimensional embeddings $Z\in\mathbb{R}^{|V|\times 28}$ obtained from the second GCN layer of the encoder.

Given two nodes $i$ and $j$, let $h_i,h_j\in\mathbb{R}^{28}$ denote their embeddings from $Z$. The decoder proceeds in three steps.

\textbf{Element-wise multiplication:} The embeddings are combined through element-wise multiplication:
\begin{equation}
e=h_i\odot h_j
\label{eq:pairwise-composition}
\end{equation}
where $\odot$ denotes the Hadamard product. This operation captures feature-level interactions between the two nodes.

\textbf{Summation:} The resulting interaction vector $e$ is reduced to a scalar score by summing over all dimensions:
\begin{equation}
s=\sum_{k=1}^{28} e_k
\label{eq:score-sum}
\end{equation}
This step aggregates the pairwise feature interactions into a single measure of similarity or compatibility.

\textbf{Sigmoid activation:} Finally, the scalar score is passed through a sigmoid function:
\begin{equation}
z_{ij}=\sigma(s)=\frac{1}{1+\exp(-s)},
\label{eq:sigmoid-prob}
\end{equation}
producing an output in the range $(0,1)$. This value can be interpreted as the probability of an edge existing between nodes $i$ and $j$.

The encoder maps high-dimensional input features into compact latent embeddings, while the decoder transforms these embeddings into probabilistic predictions of node--node relationships. Together, they form an end-to-end framework for learning graph-structured associations.

\subsubsection{Training Objective and Optimization}
We formulate disease--gene association prediction as a binary edge classification task. For each candidate pair $(u,v)$, the decoder outputs a logit score $s(u,v)$, which is transformed into a probability $\hat{y}$ via the sigmoid function. The model is trained to minimize the binary cross-entropy (BCE) loss between $\hat{y}$ and the ground-truth label $y\in\{0,1\}$, where $y=1$ denotes a known association and $y=0$ a sampled negative.

For numerical stability, we adopt the logits-based formulation of BCE (i.e., \texttt{BCEWithLogitsLoss}), which directly takes $s(u,v)$ as input:
\begin{equation}
\ell(s,y)=w_1\,y\cdot\log\!\left(1+e^{-s}\right)+w_0(1-y)\cdot\log\!\left(1+e^{s}\right),
\label{eq:weighted-logistic-loss}
\end{equation}
where $w_0$ and $w_1$ are optional class weights for negatives and positives. In our balanced dataset, we set $w_0=w_1=1$, i.e., no weighting. Preliminary experiments with slightly higher positive weights (to reflect the open-world imbalance) did not yield significant improvements, so we retained the unweighted loss. The total loss for a mini-batch $\mathcal{B}$ is the average of $\ell(s(u,v),y)$ across all pairs.

\textbf{Optimization:} We optimize parameters using the AdamW optimizer~\cite{kingma} with learning rate 0.001. Training proceeds for a fixed number of epochs $T$ without aggressive scheduling or early stopping. The final model is selected as the checkpoint achieving the highest validation AUROC during training. All reported test results are computed at this checkpoint.

\textbf{Mini-batch training:} Since the full graph may contain millions of edges, we adopt mini-batch training with subgraph sampling. At each iteration, we sample a batch $\mathcal{B}$ of positive and negative edges and construct the induced subgraph $G_{\mathcal{B}}$ containing those edges and their immediate neighbors. The GCN encoder is then applied to $G_{\mathcal{B}}$, enabling message passing within the relevant local context. This strategy reduces memory usage and accelerates training, while ensuring that over multiple epochs the model is exposed to the entire graph. To further improve efficiency, we precompute auxiliary structures such as neighbor lists and UniRef cluster IDs. The per-batch complexity of a GCN layer with output dimension $d_{\ell}$ is $\mathcal{O}(|E_{\mathcal{B}}|\cdot d_{\ell})$, and batch sizes are chosen to fit within a single GPU.

\subsection{Evaluation Metrics}
We evaluated model performance using standard binary classification metrics: Accuracy (Acc), F1-score (F1), Precision (Prec), Recall (Rec), Area Under the ROC Curve (ROC-AUC), Area Under the Precision--Recall Curve (PR-AUC), and Specificity / True Negative Rate (Spec).

\section{Results}

\subsection{Performance Evaluation Across Different Input Graphs}
We first examine GLaDiGAtor’s predictive performance on the six constructed graph variants (Graph~1--6). Table~\ref{tab:val-perf} presents the model’s accuracy, F1, precision, recall, ROC-AUC, PR-AUC, and specificity on the validation set for each graph, and Table~\ref{tab:test-perf} shows the same for the test set.

\begin{table}[!t]
\caption{Validation performance of GLaDiGAtor across different input graphs.}
\label{tab:val-perf}
\centering
\footnotesize
\setlength{\tabcolsep}{3pt}
\renewcommand{\arraystretch}{1.15}
\begin{tabular}{c c c c c c c c}
\hline
\textbf{Graph ID} & \textbf{Acc} & \textbf{F1} & \textbf{Prec} & \textbf{Rec} & \textbf{ROC-AUC} & \textbf{PR-AUC} & \textbf{Spec} \\
\hline
1 & 0.719 & 0.741 & 0.688 & 0.800 & 0.735 & 0.688 & 0.637 \\
2 & 0.830 & 0.836 & 0.805 & 0.871 & 0.896 & 0.893 & 0.789 \\
3 & 0.890 & 0.893 & 0.875 & 0.911 & 0.953 & 0.957 & 0.869 \\
4 & 0.891 & 0.895 & 0.867 & 0.955 & 0.955 & 0.960 & 0.858 \\
5 & 0.899 & 0.902 & 0.869 & 0.938 & 0.960 & 0.966 & 0.859 \\
6 & 0.857 & 0.866 & 0.815 & 0.923 & 0.943 & 0.947 & 0.791 \\
\hline
\end{tabular}
\end{table}

\begin{table}[!t]
\caption{Test performance of GLaDiGAtor across different input graphs.}
\label{tab:test-perf}
\centering
\footnotesize
\setlength{\tabcolsep}{3pt}
\renewcommand{\arraystretch}{1.15}
\begin{tabular}{c c c c c c c c}
\hline
\textbf{Graph ID} & \textbf{Acc} & \textbf{F1} & \textbf{Prec} & \textbf{Rec} & \textbf{ROC-AUC} & \textbf{PR-AUC} & \textbf{Spec} \\
\hline
1 & 0.761 & 0.786 & 0.711 & 0.877 & 0.807 & 0.746 & 0.644 \\
2 & 0.827 & 0.840 & 0.784 & 0.900 & 0.904 & 0.900 & 0.750 \\
3 & 0.890 & 0.892 & 0.874 & 0.911 & 0.953 & 0.957 & 0.869 \\
4 & 0.888 & 0.892 & 0.867 & 0.918 & 0.954 & 0.959 & 0.860 \\
5 & 0.889 & 0.893 & 0.859 & 0.930 & 0.955 & 0.960 & 0.848 \\
6 & 0.876 & 0.886 & 0.821 & 0.962 & 0.965 & 0.967 & 0.791 \\
\hline
\end{tabular}
\renewcommand{\arraystretch}{1}
\end{table}

Overall, GLaDiGAtor achieves high scores on all graphs, but the performance trends reflect the varying data characteristics:
\begin{itemize}
  \item On the smallest and most stringent dataset (Graph~1, using only very high-confidence DisGeNET associations), the model attains a test ROC-AUC of 0.807 and PR-AUC of 0.746 with accuracy 0.761. The precisions and specificity on Graph~1 is relatively low (0.711 and 0.644), indicating some false positives, likely due to the limited number of associations making negatives harder to distinguish.
  \item As the graph becomes more densely connected and includes lower-confidence but more comprehensive associations (Graphs~2, 3, 4), performance improves markedly. By Graph~3 and Graph~4, GLaDiGAtor reaches ROC-AUC $\approx 0.95$ and PR-AUC $\approx 0.96$ on test data. These larger graphs provide a richer context, enabling the model to generalise better.
  \item Graph~5 (DisGeNET curated subset) yields performance comparable to Graph~4, with test ROC-AUC $\sim 0.955$ and PR-AUC $\sim 0.960$. It is also important to note that Graph~5 is a subset of Graph~4 focusing on high-quality associations.
  \item Graph~6, derived from the OGB biokg dataset, also achieves strong results (ROC-AUC $\sim 0.965$, PR-AUC $\sim 0.967$), with slightly lower precision and specificity values compared to Graphs 3, 4 and 5. This indicates that GLaDiGAtor can transfer to a different source of associations (here, protein--disease edges from a knowledge graph) with minimal loss in performance.
  \item Overall, GLaDiGAtor demonstrates strong predictive performance under a stringent train--validation--test partitioning strategy stratified by 50\% protein sequence similarity.
\end{itemize}

\subsection{Comparison with Baseline Methods}
We compared GLaDiGAtor to 13 baseline models on the DisGeNET curated GDAs (Graph~5) dataset, to benchmark its performance against existing approaches. The baseline model performance scores are directly obtained from the SkipGNN~\cite{huang}, HOGCN~\cite{kc} and ResMGCN~\cite{yin} studies, since re-training and testing the models might lead to a decrease in their performance measurements due to using non-optimal hyperparameters.

Table~\ref{tab:baseline-comp} summarizes the ROC-AUC and PR-AUC for each method. GLaDiGAtor achieved a ROC-AUC of 0.950 and PR-AUC of 0.956, which is the highest among all methods evaluated. The next best model (HOGCN, a GCN variant for heterogeneous graphs) reached ROC-AUC 0.936 and PR-AUC 0.941, slightly below our model. Other strong performers included ResMGCN (0.925 ROC-AUC) and MixHop (0.916). Traditional network embedding approaches like node2vec or DeepWalk showed lower performance (ROC-AUC $\sim$0.83) in this task, likely because they do not incorporate rich biological features. Simpler heuristics (L3, spectral clustering) and basic GCN/VGAE models also underperformed relative to our approach. These results demonstrate that GLaDiGAtor outperforms the state-of-the-art in disease--gene association prediction.

The performance gain in GLaDiGAtor can be attributed to our model’s ability to integrate heterogeneous data (multiple edge types and node features) in a unified framework. In contrast, many baselines use either homogeneous networks or limited feature information. For example, SkipGNN and HOGCN incorporate graph structure but lack the multimodal feature integration that GLaDiGAtor has with ProtT5 and BioBERT embeddings. Our results suggest that combining graph topology with rich biological context (via pretrained embeddings) yields a more powerful predictor.

It is important to note that there are discrepancies between GLaDiGAtor's performance results in Table~\ref{tab:test-perf} and Table~\ref{tab:baseline-comp} due to different splitting ratios. In Table~\ref{tab:test-perf}, the splitting ratio is 8:1:1, whereas in Table~\ref{tab:baseline-comp}, the splitting ratio is 7:1:2, as described in the SkipGNN study~\cite{huang}.

\begin{table}[!t]
\caption{Performance comparison with baseline models on the DisGeneNET curated GDAs benchmark (Graph 5). The ranking of methods is based on ROC-AUC.}
\label{tab:baseline-comp}
\centering
\footnotesize
\setlength{\tabcolsep}{3pt}
\renewcommand{\arraystretch}{1.15}
\begin{tabular}{l c c c}
\hline
\textbf{Method} & \textbf{ROC-AUC} & \textbf{PR-AUC} & \textbf{Ranking} \\
\hline
GLaDiGAtor & 0.950 & 0.956 & 1 \\
HOGCN & 0.936 & 0.941 & 2 \\
ResMGCN & 0.925 & 0.935 & 3 \\
MixHop & 0.916 & 0.912 & 4 \\
SkipGNN & 0.912 & 0.915 & 5 \\
struc2vec & 0.909 & 0.910 & 6 \\
GCN & 0.906 & 0.909 & 7 \\
GIN & 0.900 & 0.916 & 8 \\
JK-Net & 0.898 & 0.891 & 9 \\
VGAE & 0.873 & 0.902 & 10 \\
SC & 0.863 & 0.905 & 11 \\
node2vec & 0.834 & 0.828 & 12 \\
L3 & 0.832 & 0.899 & 13 \\
DeepWalk & 0.832 & 0.827 & 14 \\
\hline
\end{tabular}
\renewcommand{\arraystretch}{1}
\end{table}

\subsection{Ablation Study}
We conducted ablation experiments to assess the contribution of key design components in GLaDiGAtor. In each ablation, we modified one aspect of the model or data processing and observed the impact on performance (using Graph~2 as a representative setting). Validation and test scores are provided in Table~\ref{tab:ablation}.
\begin{itemize}
  \item \textbf{Embedding aggregation strategy:} We tried an alternative padding alignment for combining ProtT5 and BioBERT embeddings (e.g., appending 256 zeros to the terminus of the disease embeddings, making 1024-D in total, and no addition to protein embeddings, which is already 1024-D). We found that the alternative padding scheme led to a slight drop in performance (Table~\ref{tab:ablation}) and less stable training. Our default approach preserves the full information from both modalities in a consistent orientation.
  \item \textbf{Negative sampling strategy:} We experimented with sampling all types of edges as negatives (including gene--gene or disease--disease false edges) versus our constrained approach (only gene--disease negatives, in the default model). Allowing arbitrary false edges hurts performance (Table~\ref{tab:ablation}), whereas the constrained negative sampling improves stability and maintains focus on the biologically relevant task.
\end{itemize}
In summary, the ablation results indicate that removing GLaDiGAtor’s default modules leads to consistent performance degradation, most notably in accuracy and F1 score. The only metric that remains comparatively stable under ablation is the test ROC-AUC, suggesting that the full configuration is primarily responsible for improvements in classification quality at the operating point rather than changes in overall ranking performance.

\begin{table}[H]
\caption{Ablation study performance comparison on the validation and test datasets.}
\label{tab:ablation}
\centering
\footnotesize
\setlength{\tabcolsep}{2pt}
\renewcommand{\arraystretch}{1.1}
\begin{tabular}{>{\centering\arraybackslash}m{1.6cm} >{\centering\arraybackslash}m{1.7cm} c c c c c c}
\hline
\textbf{\raisebox{-2.8ex}{\shortstack{Embedding\\aggregation}}} & \textbf{\raisebox{-2.8ex}{\shortstack{Negative\\edge gen.}}} & \textbf{\raisebox{-1.8ex}{\shortstack{Split}}} & \textbf{\raisebox{-1.8ex}{\shortstack{Acc}}} & \textbf{\raisebox{-1.8ex}{\shortstack{F1}}} & \textbf{\raisebox{-1.8ex}{\shortstack{Prec}}} & \textbf{\raisebox{-1.8ex}{\shortstack{Rec}}} & \textbf{\raisebox{-1.8ex}{\shortstack{ROC-AUC}}} \\
\specialrule{0.4pt}{2.5pt}{2pt}
\multirow{2}{*}{\rule{0pt}{2.6ex}N$^{*}$} & \multirow{2}{*}{\rule{0pt}{2.6ex}Y$^{*}$} & Val. & 0.823 & 0.826 & 0.810 & 0.843 & 0.895 \\
& & Test & 0.823 & 0.835 & 0.783 & 0.893 & 0.916 \\
\multirow{2}{*}{Y} & \multirow{2}{*}{N} & Val. & 0.739 & 0.782 & 0.671 & 0.938 & 0.870 \\
& & Test & 0.750 & 0.800 & 0.675 & 0.974 & 0.925 \\
\multirow{2}{*}{Y} & \multirow{2}{*}{Y} & Val. & 0.830 & 0.836 & 0.805 & 0.871 & 0.896 \\
& & Test & 0.827 & 0.840 & 0.784 & 0.900 & 0.904 \\
\hline
\end{tabular}
\\[2pt]
{\footnotesize $^{*}$N: no, Y: yes. Our default model is given at the bottom.}
\end{table}
\FloatBarrier

\subsection{Case Study Analysis}
To assess the biological relevance of GLaDiGAtor’s predictions, we conducted literature validation of selected gene--disease associations derived from Graph~1 (GDA score $\ge 0.9$). In this analysis, both positive and negative predictions are evaluated. For each selected case, GLaDiGAtor’s predictions and database cross-references are provided in Table~\ref{tab:case-study}.

\begin{table}[H]
\caption{Case study gene--disease association examples (Yes: the corresponding gene-disease asssociation exists according to DisGeNET -constituting the training dataset-, GLaDiGAtor or the literature; No: otherwise).}
\label{tab:case-study}
\centering
\footnotesize
\setlength{\tabcolsep}{3pt}
\renewcommand{\arraystretch}{1.15}
\begin{tabular}{l >{\raggedright\arraybackslash}p{2.4cm} c c c}
\hline
\textbf{Gene Symbol} & \textbf{Disease Name} & \textbf{DisGeNET} & \textbf{GLaDiGAtor} & \textbf{Literature} \\
\hline
ABCA1 & Tangier Disease & Yes & Yes & Yes \\
ABCA4 & Stargardt Disease & Yes & Yes & Yes \\
ACADM & MCADD & Yes & Yes & Yes \\
\hline
FANCA & Polycystic Kidney Disease I & No & No & No \\
PINK1 & Herlitz Disease & No & No & No \\
NPR2 & PHARC Syndrome & No & No & No \\
\hline
PADI4 & Rheumatoid Arthritis & No & Yes & Yes \\
GATA2 & Emberger Syndrome & No & Yes & Yes \\
SETBP1 & Schinzel--Giedion Syndrome & No & Yes & Yes \\
\hline
\end{tabular}
\renewcommand{\arraystretch}{1}
\end{table}

\begin{itemize}
  \item \textbf{Known associations rediscovered:} GLaDiGAtor correctly predicted established gene--disease relationships (Table~\ref{tab:case-study}). For instance, (i) the model linked ABCA1 with Tangier disease (TD), a rare autosomal recessive disorder marked by defective HDL metabolism~\cite{barbosa,medgen_tangier}. Mutations in ABCA1 impair cholesterol efflux, leading to HDL deficiency, peripheral cholesterol accumulation, and increased cardiovascular risk~\cite{quazi}. (ii) GLaDiGAtor identified the link between ABCA4 and Stargardt disease (STGD), an inherited retinal degeneration. The gene encodes a photoreceptor transporter vital for vitamin A cycle homeostasis. Mutations cause toxic lipofuscin buildup in the retinal pigment epithelium, leading to central vision loss~\cite{molday2018,shroyer,medgen_stargardt}. (iii) GLaDiGAtor correctly predicted the association between ACADM and Medium-chain Acyl-CoA Dehydrogenase Deficiency (MCADD). Mutations impede mitochondrial fatty acid oxidation, causing metabolite buildup and symptoms like hypoglycemia and seizures~\cite{hara,ventura}.

  \item \textbf{Correct rejection of false links:} GLaDiGAtor also correctly rejected several gene--disease pairings that might seem plausible but are not supported biologically (Table~\ref{tab:case-study}). For example, (i) FANCA and Polycystic Kidney Disease I (PKD1): GLaDiGAtor correctly rejected the association between FANCA and PKD1. While FANCA mutations cause Fanconi anemia (DNA repair defect), PKD1 stems from PKD1 gene mutations affecting renal cell architecture~\cite{peake,medgen_pkd1}. (ii) GLaDiGAtor predicted no association between PINK1 and Herlitz disease. PINK1 mutations are linked to early-onset Parkinson’s disease and mitochondrial dysfunction~\cite{valente2004pink1}, while Herlitz disease is an autoimmune blistering disorder targeting type VII collagen~\cite{medgen_herlitz}. (iii) GLaDiGAtor correctly identified no link between NPR2 and PHARC syndrome. PHARC arises from ABHD12 mutations and affects neurological and sensory functions~\cite{omim_pharc}. NPR2 encodes a membrane receptor with guanylate cyclase activity (GC-B/NPR-B) that binds C-type natriuretic peptide (CNP), thereby catalyzing cGMP production~\cite{uniprot_npr2}. This function is unrelated to PHARC pathophysiology.

  \item \textbf{Novel predictions supported by literature:} Importantly, GLaDiGAtor proposed several gene--disease associations that were not present in DisGeNET (at the time of our analysis) but for which we found independent literature evidence, suggesting they are valid novel findings (Table~\ref{tab:case-study}). Three compelling examples are: (i) GLaDiGAtor accurately predicted the association between PADI4 and Rheumatoid Arthritis (RA). PADI4 encodes a protein citrullination enzyme that produces autoantigens implicated in RA. Multiple studies reported a positive association between PADI4 variants and RA susceptibility~\cite{iwamoto2006padi4}. (ii) GLaDiGAtor correctly linked GATA2 to Emberger syndrome, a condition combining immunodeficiency, lymphedema, and MDS/AML. Haploinsufficiency of GATA2 disrupts hematopoietic and lymphatic regulation~\cite{blood_gata2_protean,ostergaard2011gataa2}. (iii) GLaDiGAtor predicted the association between SETBP1 and Schinzel--Giedion syndrome (SGS), which is caused by recurrent de novo gain-of-function SETBP1 variants~\cite{acuna} and is supported by transcriptomic dysregulation in a Setbp1$^{S858R}$ mouse model~\cite{setbp1_whitlock}. None of these gene--disease pairs were indexed in DisGeNET at the time of our analysis, yet our model predicted them as positive associations (which are counted as false positives in our performance evaluation). However, subsequent literature review revealed studies supporting each link.
\end{itemize}

\section{Discussion}
In this work, we introduced GLaDiGAtor, a machine learning framework that integrates graph representation learning with contextual biomedical embeddings to predict disease--gene associations. The motivation for GLaDiGAtor arises from the growing need for scalable, accurate methods to map the genetic landscape of diseases, as traditional curation cannot keep up with data growth. GLaDiGAtor performs multi-relation graph convolution over gene--gene, disease--disease, and gene--disease edges, and augments nodes with ProtT5 (protein) and BioBERT (disease-text) embeddings.

Model comparisons with state-of-the-art models (e.g., HOGCN~\cite{kc}, ResMGCN~\cite{yin}, SkipGNN~\cite{huang}) demonstrate GLaDiGAtor’s superior or competitive performance in disease--gene association prediction. Ablation studies confirm the importance of effective embedding aggregation and constrained negative edge generation strategies. The case study analysis highlights GLaDiGAtor’s robust ability to prioritise biologically meaningful gene--disease associations. Among the nine evaluated cases derived from Graph~1 (GDA score $\ge 0.9$), the model consistently aligned with known ground truth, correctly identifying both true and false associations.

\subsection{Limitations}
Despite the promising outcomes, our approach has several limitations that must be acknowledged:
\begin{itemize}
  \item \textbf{Data dependence:} The model’s accuracy is influenced by the quality, completeness, and biases of the input databases. If important gene--disease links are missing or if spurious associations are present in the data, the model’s learning and outputs will be affected. GLaDiGAtor inherits any biases in DisGeNET or other sources (e.g., well-studied genes might appear overly important).
  \item \textbf{Computational complexity:} Training a GCN on large heterogeneous graphs is resource-intensive. GLaDiGAtor’s scalability may be constrained by memory and time requirements, especially as we move to graphs with millions of nodes or edges. Techniques such as mini-batching and subgraph sampling help, but very large graphs or those requiring higher-dimensional embeddings could pose challenges.
  \item \textbf{Limited interpretability of predictions:} Although the model can rank candidate gene--disease links, it does not explicitly provide mechanistic rationales (e.g., which specific neighbors, relation types, or embedding dimensions drive an individual prediction). Additional explainability analyses would be needed to support clinical or experimental decision-making.
\end{itemize}

\subsection{Future Work}
Future work should explore improved embedding algorithms, alternative GCN architectures, dynamic graph learning, and expanded datasets to refine accuracy and increase real-world applicability. Extending GLaDiGAtor toward multimodal association prediction represents a promising direction for advancing the field of biological/biomedical association prediction.

\section{Conclusion}
GLaDiGAtor provides a scalable, graph-driven approach for prioritising gene candidates for genetic diseases by combining heterogeneous network structure with contextual biomedical embeddings. The proposed framework achieves strong predictive performance, making it a practical tool for guiding follow-up biomedical investigations.

\bibliographystyle{IEEEtran}
\bibliography{refs}

@article{ref3,
  author  = {Topol, Eric},
  title   = {High-performance medicine: the convergence of human and artificial intelligence},
  journal = {Nature Medicine},
  year    = {2019},
  volume  = {25},
  pages   = {44--56}
}

@article{ref4,
  author  = {Botstein, David and Risch, Neil},
  title   = {Discovering genotypes underlying human phenotypes: past successes and future approaches},
  journal = {Nature Genetics},
  year    = {2003},
  volume  = {33},
  number  = {Suppl},
  pages   = {228--237}
}

@article{ref5,
  author  = {Richards, Sue and Aziz, Nazneen and Bale, Sherri and Bick, David and Das, Soma and Gastier-Foster, Julie and Grody, Wayne W. and Hegde, Madhuri and Lyon, Elaine and Spector, Elaine and Voelkerding, Karl and Rehm, Heidi L.},
  title   = {Standards and guidelines for the interpretation of sequence variants: a joint consensus recommendation of the {American College of Medical Genetics and Genomics} and the {Association for Molecular Pathology}},
  journal = {Genetics in Medicine},
  year    = {2015},
  volume  = {17},
  number  = {5},
  pages   = {405--424}
}

@article{ref1,
  author  = {Pi{\~n}ero, Janet and Ram{\'i}rez-Anguita, Juan M. and Sauch-Pitarch, Jordi and Ronzano, Francesco and Centeno, Emilio and Sanz, Ferran and Furlong, Laura I.},
  title   = {The {DisGeNET} knowledge platform for disease genomics: 2019 update},
  journal = {Nucleic Acids Research},
  year    = {2020},
  volume  = {48},
  number  = {D1},
  pages   = {D845--D855}
}

@article{ref7,
  author  = {He, Meijin and Huang, Chuan and Liu, Bo and Wang, Yaxuan and Li, Jun},
  title   = {Factor graph-aggregated heterogeneous network embedding for disease-gene association prediction},
  journal = {BMC Bioinformatics},
  year    = {2021},
  volume  = {22},
  pages   = {165}
}

@misc{ref23,
  author = {{DisGeNET}},
  title  = {{DisGeNET} database information},
  note   = {Accessed: 2024-02-27}
}

@article{ref25,
  author  = {Lee, Jinhyuk and Yoon, Wonjin and Kim, Sungdong and Kim, Donghyeon and Kim, Sunkyu and So, Chan and Kang, Jaewoo},
  title   = {{BioBERT}: a pre-trained biomedical language representation model for biomedical text mining},
  journal = {Bioinformatics},
  year    = {2020},
  volume  = {36},
  pages   = {1234--1240}
}

@article{dogan2021druidom,
  author  = {Do{\u{g}}an, Tunca and G{\"u}zelcan, E. A. and Baumann, M. and Koyas, A. and Ata{\c{s}}, H. and Baxendale, I. R. and Martin, M. and Cetin-Atalay, R.},
  title   = {Protein domain-based prediction of drug/compound--target interactions and experimental validation on LIM kinases},
  journal = {PLOS Computational Biology},
  year    = {2021},
  volume  = {17},
  number  = {11},
  pages   = {e1009171}
}

@article{huang,
  author  = {Huang, K. and Xiao, C. and Glass, L. and Zitnik, M. and Sun, J.},
  title   = {SkipGNN: predicting molecular interactions with skip-graph networks},
  journal = {Scientific Reports},
  year    = {2020},
  volume  = {10}
}

@article{kc,
  author  = {KC, K. and Li, R. and Cui, F. and Haake, A.},
  title   = {Predicting Biomedical Interactions With Higher-Order Graph Convolutional Networks},
  journal = {IEEE/ACM Transactions on Computational Biology and Bioinformatics},
  year    = {2022},
  volume  = {19},
  pages   = {676--687}
}

@misc{yin,
  author = {Yin, Z.},
  title  = {ResMGCN: Residual Message Graph Convolution Network for Fast Biomedical Interactions Discovering},
  note   = {arXiv preprint arXiv:2311.07632},
  year   = {2023}
}

@misc{kipf1,
  author = {Kipf, T. N. and Welling, M.},
  title  = {Semi-supervised classification with graph convolutional networks},
  note   = {arXiv preprint arXiv:1609.02907},
  year   = {2016}
}

@misc{vgae,
  author = {Kipf, T. N. and Welling, M.},
  title  = {Variational graph auto-encoders},
  note   = {arXiv preprint arXiv:1611.07308},
  year   = {2016}
}

@misc{xu1,
  author = {Xu, K. and Hu, W. and Leskovec, J. and Jegelka, S.},
  title  = {How powerful are graph neural networks?},
  note   = {arXiv preprint arXiv:1810.00826},
  year   = {2018}
}

@inproceedings{xu2,
  author    = {Xu, K. and Li, C. and Tian, Y. and Sonobe, T. and Kawarabayashi, K. and Jegelka, S.},
  title     = {Representation learning on graphs with jumping knowledge networks},
  booktitle = {International Conference on Machine Learning},
  year      = {2018},
  pages     = {5453--5462}
}

@article{unlu2025,
  author  = {{\"U}nl{\"u}, A. and {\c{C}}evrim, E. and Yi{\u{g}}it, M. G. and Sar{\i}g{\"u}n, A. and {\c{C}}elikbilek, H. and Bayram, O. and others and Do{\u{g}}an, T.},
  title   = {Target-specific de novo design of drug candidate molecules with graph-transformer-based generative adversarial networks},
  journal = {Nature Machine Intelligence},
  year    = {2025},
  volume  = {7},
  number  = {9},
  pages   = {1524--1540}
}

@inproceedings{abuelhaija,
  author    = {Abu-El-Haija, S. and Perozzi, B. and Kapoor, A. and Alipourfard, N. and Lerman, K. and Harutyunyan, H. and Ver Steeg, G. and Galstyan, A.},
  title     = {MixHop: Higher-order graph convolutional architectures via sparsified neighborhood mixing},
  booktitle = {International Conference on Machine Learning},
  year      = {2019},
  pages     = {21--29}
}

@article{kovacs,
  author  = {Kovacs, I. and Luck, K. and Spirohn, K. and Wang, Y. and Pollis, C. and Bian, W. and Kim, D. and Kishore, N. and Hao, T. and others},
  title   = {Network-based prediction of protein interactions},
  journal = {Nature Communications},
  year    = {2019},
  volume  = {10},
  pages   = {1240}
}

@article{tang,
  author  = {Tang, L. and Liu, H.},
  title   = {Leveraging social media networks for classification},
  journal = {Data Mining and Knowledge Discovery},
  year    = {2011},
  volume  = {23},
  pages   = {447--478}
}

@inproceedings{perozzi,
  author    = {Perozzi, B. and Al-Rfou, R. and Skiena, S.},
  title     = {DeepWalk: Online learning of social representations},
  booktitle = {Proceedings of the 20th ACM SIGKDD International Conference on Knowledge Discovery and Data Mining},
  year      = {2014},
  pages     = {701--710}
}

@inproceedings{grover,
  author    = {Grover, A. and Leskovec, J.},
  title     = {node2vec: Scalable feature learning for networks},
  booktitle = {Proceedings of the 22nd ACM SIGKDD International Conference on Knowledge Discovery and Data Mining},
  year      = {2016},
  pages     = {855--864}
}

@inproceedings{ribeiro,
  author    = {Ribeiro, L. and Saverese, P. and Figueiredo, D.},
  title     = {struc2vec: Learning node representations from structural identity},
  booktitle = {Proceedings of the 23rd ACM SIGKDD International Conference on Knowledge Discovery and Data Mining},
  year      = {2017},
  pages     = {385--394}
}

@article{elnaggar,
  author  = {Elnaggar, A. and Heinzinger, M. and Dallago, C. and Rehawi, G. and Wang, Y. and Jones, L. and others and Rost, B.},
  title   = {ProtTrans: Toward understanding the language of life through self-supervised learning},
  journal = {IEEE Transactions on Pattern Analysis and Machine Intelligence},
  year    = {2021},
  volume  = {44},
  number  = {10},
  pages   = {7112--7127}
}

@article{oughtred,
  author  = {Oughtred, R. and Stark, C. and Breitkreutz, B. and Rust, J. and Boucher, L. and Chang, C. and Kolas, N. and O'Donnell, L. and Leung, G. and McAdam, R. and others},
  title   = {The BioGRID interaction database: 2019 update},
  journal = {Nucleic Acids Research},
  year    = {2019},
  volume  = {47},
  pages   = {D529--D541}
}

@misc{huogb,
  author = {Hu, W. and Fey, M. and Zitnik, M. and Dong, Y. and Ren, H. and Liu, B. and Catasta, M. and Leskovec, J.},
  title  = {Open Graph Benchmark: Datasets for Machine Learning on Graphs},
  year   = {2021}
}

@techreport{hagberg,
  author      = {Hagberg, A. and Swart, P. and Chult, D.},
  title       = {Exploring network structure, dynamics, and function using NetworkX},
  institution = {Los Alamos National Laboratory (LANL)},
  year        = {2008}
}

@misc{fey,
  author = {Fey, M. and Lenssen, J.},
  title  = {Fast Graph Representation Learning with PyTorch Geometric},
  note   = {ICLR Workshop on Representation Learning on Graphs and Manifolds},
  year   = {2019}
}

@article{suzek,
  author  = {Suzek, B. E. and Wang, Y. and Huang, H. and McGarvey, P. and Wu, C. and UniProt Consortium},
  title   = {UniRef clusters: a comprehensive and scalable alternative for improving sequence similarity searches},
  journal = {Bioinformatics},
  year    = {2015},
  volume  = {31},
  pages   = {926--932}
}

@misc{kingma,
  author = {Kingma, D. P. and Ba, J.},
  title  = {Adam: A method for stochastic optimization},
  note   = {arXiv preprint arXiv:1412.6980},
  year   = {2014}
}

@article{quazi,
  author  = {Quazi, F. and Molday, R.},
  title   = {Differential phospholipid substrates and directional transport by ATP-binding cassette proteins ABCA1, ABCA7, and ABCA4 and disease-causing mutants},
  journal = {Journal of Biological Chemistry},
  year    = {2013},
  volume  = {288},
  pages   = {34414--34426}
}

@article{molday2018,
  author  = {Molday, L. and Wahl, D. and Sarunic, M. and Molday, R.},
  title   = {Localization and functional characterization of the p.Asn965Ser (N965S) ABCA4 variant in mice reveal pathogenic mechanisms underlying Stargardt macular degeneration},
  journal = {Human Molecular Genetics},
  year    = {2018},
  volume  = {27},
  pages   = {295--306}
}

@article{shroyer,
  author  = {Shroyer, N. and Lewis, R. and Yatsenko, A. and Lupski, J.},
  title   = {Null missense ABCR (ABCA4) mutations in a family with Stargardt disease and retinitis pigmentosa},
  journal = {Investigative Ophthalmology \& Visual Science},
  year    = {2001},
  volume  = {42},
  pages   = {2757--2761}
}

@article{barbosa,
  author  = {Barbosa-Gouveia, S. and Fernandez-Crespo, S. and Lazare-Iglesias, H. and Gonzalez-Quintela, A. and Vazquez-Agra, N. and Hermida-Ameijeiras, A.},
  title   = {Association of a Novel Homozygous Variant in ABCA1 Gene with Tangier Disease},
  journal = {Journal of Clinical Medicine},
  year    = {2023},
  volume  = {12}
}

@article{hara,
  author  = {Hara, K. and Tajima, G. and Okada, S. and Tsumura, M. and Kagawa, R. and Shirao, K. and Ohno, Y. and Yasunaga, S. and Ohtsubo, M. and Hata, I. and Sakura, N. and Shigematsu, Y. and Takihara, Y. and Kobayashi, M.},
  title   = {Significance of ACADM mutations identified through newborn screening of MCAD deficiency in Japan},
  journal = {Molecular Genetics and Metabolism},
  year    = {2016},
  volume  = {118},
  pages   = {9--14}
}

@article{ventura,
  author  = {Ventura, F. and Leandro, P. and Luz, A. and Rivera, I. and Silva, M. and Ramos, R. and Rocha, H. and Lopes, A. and Fonseca, H. and Gaspar, A. and Diogo, L. and Martins, E. and Leao-Teles, E. and Vilarinho, L. and Almeida, I.},
  title   = {Retrospective study of the medium-chain acyl-CoA dehydrogenase deficiency in Portugal},
  journal = {Clinical Genetics},
  year    = {2014},
  volume  = {85},
  pages   = {555--561}
}

@article{peake,
  author  = {Peake, J. D. and Noguchi, E.},
  title   = {Fanconi anemia: current insights regarding epidemiology, cancer, and DNA repair},
  journal = {Human Genetics},
  year    = {2022},
  volume  = {141},
  number  = {12},
  pages   = {1811--1836}
}

@misc{medgen_herlitz,
  author       = {{NCBI MedGen}},
  title        = {Junctional epidermolysis bullosa, Herlitz type},
  howpublished = {National Center for Biotechnology Information (NCBI)},
  year         = {2024},
  url          = {https://www.ncbi.nlm.nih.gov/medgen/86898},
  note         = {Accessed: 2024-02-27}
}

@article{valente2004pink1,
  author  = {Valente, E. M. and Abou-Sleiman, P. M. and Caputo, V. and Muqit, M. M. K. and Harvey, K. and Gispert, S. and Ali, Z. and Del Turco, D. and Bentivoglio, A. R. and Healy, D. G. and Albanese, A. and Nussbaum, R. and Gonzalez-Maldonado, R. and Deller, T. and Salvi, S. and Cortelli, P. and Gilks, W. P. and Latchman, D. S. and Harvey, R. J. and Dallapiccola, B. and Auburger, G. and Wood, N. W.},
  title   = {Hereditary early-onset Parkinson's disease caused by mutations in {PINK1}},
  journal = {Science},
  year    = {2004},
  volume  = {304},
  number  = {5674},
  pages   = {1158--1160}
}

@article{setbp1_whitlock,
  author  = {Whitlock, J. H. and others},
  title   = {Cell-type-specific gene expression and regulation in the cerebral cortex and kidney of atypical Setbp1 S858R Schinzel--Giedion Syndrome mice},
  journal = {Journal of Cellular and Molecular Medicine},
  year    = {2023},
  volume  = {27},
  number  = {22},
  pages   = {3565--3577}
}

@article{acuna,
  author  = {Acuna-Hidalgo, R. and others},
  title   = {Overlapping SETBP1 gain-of-function mutations in Schinzel--Giedion syndrome and hematologic malignancies},
  journal = {PLOS Genetics},
  year    = {2017},
  volume  = {13},
  number  = {3},
  pages   = {e1006683}
}

@misc{medgen_tangier,
  author       = {{NCBI MedGen}},
  title        = {Tangier disease},
  howpublished = {National Center for Biotechnology Information (NCBI)},
  year         = {2024},
  url          = {https://www.ncbi.nlm.nih.gov/medgen/52644},
  note         = {Accessed: 2024-02-27}
}

@misc{medgen_stargardt,
  author       = {{NCBI MedGen}},
  title        = {Stargardt disease},
  howpublished = {National Center for Biotechnology Information (NCBI)},
  year         = {2024},
  url          = {https://www.ncbi.nlm.nih.gov/medgen/75734},
  note         = {Accessed: 2024-02-27}
}

@misc{medgen_pkd1,
  author       = {{NCBI MedGen}},
  title        = {Polycystic kidney disease, adult type},
  howpublished = {National Center for Biotechnology Information (NCBI)},
  year         = {2024},
  url          = {https://www.ncbi.nlm.nih.gov/medgen/461191},
  note         = {Accessed: 2024-02-27}
}

@article{iwamoto2006padi4,
  author  = {Iwamoto, T. and Ikari, K. and Nakamura, T. and Kuwahara, M. and Toyama, Y. and Tomatsu, T. and others and Kamatani, N.},
  title   = {Association between {PADI4} and rheumatoid arthritis: a meta-analysis},
  journal = {Rheumatology},
  year    = {2006},
  volume  = {45},
  number  = {7},
  pages   = {804--807}
}

@article{ostergaard2011gataa2,
  author  = {Ostergaard, P. and Simpson, M. A. and Connell, F. C. and Steward, C. G. and Brice, G. and Woollard, W. J. and others and Mansour, S.},
  title   = {Mutations in {GATA2} cause primary lymphedema associated with a predisposition to acute myeloid leukemia (Emberger syndrome)},
  journal = {Nature Genetics},
  year    = {2011},
  volume  = {43},
  number  = {10},
  pages   = {929--931}
}

@article{blood_gata2_protean,
  author       = {Spinner, Michael A. and Sanchez, Liliana A. and Hsu, Anna P. and Shaw, Pamela A. and Zerbe, Christina S. and Calvo, Katherine R. and Arthur, David C. and Gu, Wei and Gould, Chris M. and Brewer, Chad C. and Cowen, Edward W. and Freeman, Alexandra F. and Olivier, Kenneth N. and Uzel, Gulbu and Zelazny, Adrian M. and Daub, Jeffrey R. and Spalding, Christopher D. and Claypool, Rebecca J. and Giri, Neelam K. and Alter, Blanche P. and Mace, Emily M. and Orange, Jordan S. and Cuellar-Rodriguez, Jenifer and Hickstein, Dennis D. and Holland, Steven M.},
  title        = {{GATA2} deficiency: a protean disorder of hematopoiesis, lymphatics, and immunity},
  journal = {Blood},
  year    = {2014},
  volume  = {123},
  number  = {6},
  pages   = {809--821},
  doi     = {10.1182/blood-2013-07-515528}
}

@misc{omim_pharc,
  author       = {{OMIM}},
  title        = {Polyneuropathy, Hearing Loss, Ataxia, Retinitis Pigmentosa, and Cataract},
  howpublished = {Online Mendelian Inheritance in Man (OMIM)},
  year         = {2024},
  url          = {https://www.omim.org/entry/612674},
  note         = {Accessed: 2024-02-27}
}

@misc{uniprot_npr2,
  author       = {{UniProt Consortium}},
  title        = {Natriuretic peptide receptor 2 ({Homo sapiens}) (NPR2) (UniProtKB: P20594)},
  howpublished = {UniProt},
  year         = {2024},
  url          = {https://www.uniprot.org/uniprotkb/P20594/entry},
  note         = {Accessed: 2024-02-27}
}

\end{document}